\documentclass[sigconf, screen, nonacm]{acmart}
\AtBeginDocument{%
  }

\setcopyright{acmlicensed}
\copyrightyear{2025}
\acmYear{2025}
\acmDOI{XXXXXXX.XXXXXXX}
\acmConference[ACM MM]{The 33rd ACM International Conference on Multimedia}{October 27--31, 2025}{Dublin, Ireland}
\acmISBN{978-1-4503-XXXX-X/2025/04}

\DeclareMathOperator*{\argmax}{arg\,max}

\definecolor{Mycolor}{HTML}{0066cc}
\newif\ifshowappendix
\showappendixtrue % Set to true to show, false to hide

\begin{document}

\title[Retrieval-based Explainable Multimodal Evidence-guided Modeling for Brain Evaluation and Reasoning]{REMEMBER: Retrieval-based Explainable Multimodal Evidence-guided Modeling for Brain Evaluation and Reasoning in~Zero- and Few-shot Neurodegenerative Diagnosis}

\author{Duy-Cat Can}
\email{duy-cat.can@chuv.ch}
\orcid{0000-0002-6861-2893}
\affiliation{%
  % \institution{Plateforme de bio-informatique, Centre hospitalier universitaire vaudois (CHUV)}
  % \institution{Faculté de biologie et de médecine, Université de Lausanne (UNIL)}
  \institution{Lausanne University Hospital (CHUV)}
  \institution{University of Lausanne (UNIL)}
  \city{}
  \country{}
  %\city{Lausanne}
  %\state{Vaud}
  %\country{Switzerland}\\
  \institution{and Vietnam National University}
  \city{}
  \country{}
  % \city{Hanoi}
  % \country{Vietnam}
}

\author{Quang-Huy Tang}
\email{tqhuy23@apcs.fitus.edu.vn}
\orcid{0009-0005-8044-051X}
\affiliation{%
  \institution{Department of Computer Science, University of Science, Vietnam National University Ho Chi Minh City}
  \city{}
  \country{}
  %\city{Ho Chi Minh City}
  %\country{Vietnam}
}

\author{Huong Ha}
\email{htthuong@hcmiu.edu.vn}
\orcid{0000-0002-8884-8692}
\affiliation{%
  \institution{School of Biomedical Engineering, International University, Vietnam National University Ho Chi Minh City}
  \city{}
  \country{}
  %\city{Ho Chi Minh City}
  %\country{Vietnam}
}

\author{Binh T. Nguyen}
\authornotemark[1]
\email{ngtbinh@hcmus.edu.vn}
\orcid{0000-0001-5249-9702}
\affiliation{%
  \institution{
  %Department of Computer Science, 
  University of Science, Vietnam National University Ho Chi Minh City}
  \city{}
  \country{}
  %\city{Ho Chi Minh City}
  %\country{Vietnam}
}

\author{Oliver Y. Ch\'en}
\authornote{Corresponding authors.}
\email{olivery.chen@chuv.ch}
\orcid{0000-0002-5696-3127}
\affiliation{%
  % \institution{Plateforme de bio-informatique, Centre hospitalier universitaire vaudois (CHUV)}
  % \institution{Faculté de biologie et de médecine, Université de Lausanne (UNIL)}
  \institution{Lausanne University Hospital (CHUV)}
  \institution{and University of Lausanne (UNIL)}
  \city{}
  \country{}
  %\city{Lausanne}
  %\state{Vaud}
  %\country{Switzerland}
  % \vspace{2mm}
}

% \affiliation{%

\renewcommand{\shortauthors}{Can et al.}


\begin{abstract}
    Timely and accurate diagnosis of neurodegenerative disorders, such as Alzheimer's disease, is central to disease management.
    Existing deep learning models require large-scale annotated datasets and often function as ``black boxes''. Additionally, datasets in clinical practice are frequently small or unlabeled, restricting the full potential of deep learning methods.
    Here, we introduce REMEMBER -- \textbf{R}etrieval-based \textbf{E}xplainable \textbf{M}ultimodal \textbf{E}vidence-guided \textbf{M}odeling for \textbf{B}rain \textbf{E}valuation and \textbf{R}easoning -- a new machine learning framework that facilitates zero- and few-shot Alzheimer's diagnosis using brain MRI scans through a reference-based reasoning process.
    Specifically, REMEMBER first trains a contrastively aligned vision-text model using expert-annotated reference data and extends pseudo-text modalities that encode abnormality types, diagnosis labels, and composite clinical descriptions.
    Then, at inference time, REMEMBER retrieves similar, human-validated cases from a curated dataset and integrates their contextual information through a dedicated evidence encoding module and attention-based inference head.
    Such an evidence-guided design enables REMEMBER to imitate real-world clinical decision-making process by grounding predictions in retrieved imaging and textual context.
    Specifically, REMEMBER outputs diagnostic predictions alongside an interpretable report, including reference images and explanations aligned with clinical workflows.
    Experimental results demonstrate that REMEMBER achieves robust zero- and few-shot performance and offers a powerful and explainable framework to neuroimaging-based diagnosis in the real world, especially under limited data.
\end{abstract}



\begin{teaserfigure}
\centering
\includegraphics[width=\textwidth]{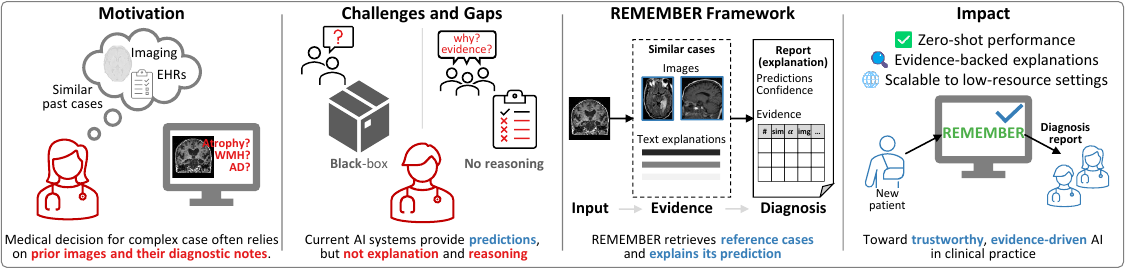}
    \caption{
        \textbf{An overview of the motivation, challenges and gaps, solution, and clinical impact for building REMEMBER.}
        \normalfont
        Given a brain scan from a new subject, REMEMBER looks at the scan, retrieves existing scans most similar to this scan from a curated dataset that has been annotated by human experts, and synthesizes their image and textual information. It outputs both disease prediction and a report with clinical explanations to support transparent, evidence-guided clinical inference. REMEMBER's machine-assisted clinical reasoning is similar to physicians making clinical decisions by comparing a new case with existing examples and their diagnostic reports.
    }
    \Description{
        The framework retrieves similar reference cases to support transparent, evidence-guided prediction and explanation.
    }
    \label{fig:teaser}
    \vspace{1em}
\end{teaserfigure}

% \received{11 April 2025}
% \received[revised]{xx xxxx 2025}
% \received[accepted]{xx xxxx 2025}

\maketitle

% ===============================================================
% SECTION INTRODUCTION
% ===============================================================
\section{Introduction}
% Motivation
Alzheimer's Disease (AD) and related neurodegenerative disorders represent one of the most urgent global health concerns with growing prevalence%and limited treatment option
% I would not say limited treatment option. People would ask what is the point of predicting it when there are not treatments. There are some treatment - although not very effective. One can say early prediction helps disease management. But in either case, it does not help to say there is limited treatment options. 
~\cite{winblad2016defeating, jack2018ninds}.
Early and accurate diagnosis is critical to timely interventions, but it is at present challenging, due in part to the subtle and heterogeneous nature of early symptoms~\cite{sirkis2022dissecting}.
One way to capture AD'S minor signs and individual-differences is to use structural brain imaging, particularly magnetic resonance imaging (MRI), as brain structural changes likely precedes symptomatic signs~\cite{frisoni2010clinical,illakiya2023automatic}.
Nevertheless, reliably interpreting neuroimaging data in practice requires years of experience and is often complicated by inter-individual variability and non-specific findings~\cite{bron2015standardized}.

% Image understanding with Deep learning
Recent years have seen deep learning's promises in automating neurodegenerative disease diagnosis using MRI scans~\cite{wen2020convolutional, li2021novel, menagadevi2024machine}.
Despite progress, existing models face two key limitations.
First, they rely heavily on large-scale labeled datasets, which are often difficult to obtain in clinical domains.
Second, deep learning models, as-of-yet, typically operate as opaque ``black-box'' systems.
Naturally, this offers little insight into the biological basis of their predictions -- a barrier to real-world adoption~\cite{tjoa2020survey}.

% Vision-language model
Recent advances in vision-language modeling, especially contrastively trained architectures such as CLIP~\cite{radford2021learning}, its biomedical adaptations such as BioMedCLIP~\cite{zhang2024biomedclip}, and domain-specific frameworks such as VisTA~\cite{can2025vista}, offer promises toward generalizable and explainable medical AI.
These models align visual inputs with textual descriptions in a shared embedding space, enabling zero-shot classification and more interpretable outputs,
without the need for task-specific supervision.
% By leveraging large-scale image-text pairings, such systems can perform classification without requiring task-specific supervision.
However, most current approaches focus on matching queries to pre-defined description and lack the key spirit of clinical decision-making, that is to provide contextualized reasoning grounded in retrieved reference cases, where physicians compare a new case with existing examples and diagnostic reports.

% Proposed method
To address these challenges, we propose \textbf{REMEMBER} (\textbf{R}etrieval-based \textbf{E}xplainable \textbf{M}ultimodal \textbf{E}vidence-guided \textbf{M}odeling for \textbf{B}rain \textbf{E}valuation and \textbf{R}easoning), a framework that enables \textbf{zero- and few-shot neurodegenerative prediction} from MRI by simulating clinician-style reasoning through multimodal retrieval.
Instead of relying solely on large-scale supervision, REMEMBER retrieves expert-annotated cases from a curated dataset and synthesizes image and textual information for evidence-guided inference.

% Contribution
The first contribution of REMEMBER is a \textbf{\textit{domain-adapted vision-language model}} that aligns radiology images with clinically meaningful textual descriptions.
Using contrastive learning, we extend existing vision-language models by incorporating diverse pseudo-text modalities derived from expert-annotated reference cases, including abnormality types, diagnostic labels, and composite descriptions.
This approach allows REMEMBER to project both image and text modalities into a shared semantic space, enabling accurate image-text matching under zero- and few-shot settings without relying on supervised labels, as done traditionally.

Second, REMEMBER introduces a \textbf{\textit{retrieval-guided inference framework}} that performs diagnosis through evidence aggregation and attention-based reasoning.
More specifically, given a query image from a new subject, REMEMBER retrieves the top-$k$ most similar cases from a reference dataset, encodes their contextual information via an Evidence Encoding Module, and integrates it using an Attention-based Inference Head.
% By doing so, REMEMBER is able to perform clinically grounded predictions by explicitly referencing to confirmed cases, thereby improving accuracy and transparency~\cite{rajpurkar2022ai}.
This enables clinically grounded predictions informed by real, validated examples, enhancing both accuracy and transparency~\cite{rajpurkar2022ai}.

Third, REMEMBER \textbf{\textit{outputs clinical explanations}} alongside its predictions.
Instead of returning a single label, it generates diagnostic reports including retrieved cases, %(existing subjects whose disease profiles are most similar to the new subject under investigation),
abnormality descriptions, and supporting evidence -- providing a transparent clinical justification for each decision endorsed by confirmed cases.
% This supports trustworthy and actionable AI in healthcare, and potentially narrows the gap between ``machine intelligence'' tools, human knowledge, and the feasibility of utilizing AI in real clinical practices~\cite{ghassemi2020review,doshi2017towards}.
This supports trustworthy and actionable AI in healthcare, bridging the gap between machine predictions and clinical reasoning~\cite{ghassemi2020review,doshi2017towards}.

% experimental and application
% To demonstrate REMEMBER's efficacy,
To evaluate REMEMBER, we test it on a hybrid dataset comprising the expert-verified MINDSet reference corpus and a public Alzheimer's dataset.
% Our results show that REMEMBER achieves good performance in both zero-shot and few-shot settings, and offers interpretable, evidence-driven outputs.
REMEMBER demonstrates strong zero-/few-shot performance while generating interpretable, evidence-driven outputs.
% Taken together, REMEMBER introduces a new clinical AI paradigm for retrieval-based, multimodal, and explainable medical intelligence that is capable of operating with little supervision when making important clinical decisions.
%in low-supervision, high-stakes environments.
Together, REMEMBER introduces a new paradigm for retrieval-based, multimodal, and explainable clinical AI, particularly suited for low-supervision scenarios in real-world healthcare.

% ===============================================================
% RELATED WORK
% ===============================================================
\section{Related Work}
\textit{\textbf{Neurodegenerative Disease Diagnosis.}}
Neuroimaging data are important for the diagnosis of neurodegenerative diseases, such as Alzheimer's disease.
An important neuroimaging modality is magnetic resonance imaging, which depicts anatomical and physiological changes of the brain~\cite{frisoni2010clinical, jack2018ninds}.
Classical MRI analysis often involve hand-crafted radiomic features and statistical models to detect atrophy or structural abnormalities~\cite{cuingnet2011automatic, kloppel2008automatic}.
Deep learning methods, such as convolutional neural networks (CNNs)~\cite{wen2020convolutional, li2021novel, kaur2024systematic, el2024novel}, have demonstrated strong performance by learning disease-related patterns directly from imaging data.
More recently, transformer-based architectures have been explored for modeling long-range spatial dependencies in volumetric scans~\cite{zhang2023transformer}.

In parallel, multimodal data analysis integrating clinical, cognitive, genetic, and biochemical data has shown promises in improving diagnostic performance~\cite{venugopalan2021multimodal}.
Nevertheless, training a multimodal often requires large-scale labeled datasets and obtaining multimodal data is both time- and resource-consuming.
% Additionally, multimodal data are usually distinct from each other, and are difficult to align and, therefore, challenging to interpret.
These limitations raise concerns for their real-world clinical deployment~\cite{ghassemi2020review}.
To address them, our work explores lightweight, explainable alternatives requiring minimal supervision.

\textit{\textbf{Vision-Language Models in Medical Imaging.}}
Recent advances in vision-language pretraining have introduced powerful models capable of aligning visual inputs with textual semantics.
For example, CLIP~\cite{radford2021learning} demonstrated the effectiveness of contrastive learning on large-scale image-text pairs, enabling zero-shot classification across a wide range of categories.

Domain-specific adaptations have extended these approaches to more complex or fine-grained medical tasks. 
GLoRIA~\cite{huang2021gloria} and MedCLIP~\cite{wang2022medclip} align radiographs with report sections for improved grounding.
BiomedCLIP~\cite{zhang2024biomedclip} adapts this strategy to biomedical imaging, achieving strong performance in retrieval and classification of chest X-rays.
VisTA~\cite{can2025vista} applies CLIP-style training to neuroimaging, aligning MRI scans with textual descriptions of abnormalities and diagnoses for zero-shot dementia classification.

While these models enable interpretable, text-driven predictions, most focus on label matching and thus lack the capacity to retrieve and integrate case-level reference evidence.
This gap motivates REMEMBER's retrieval-augmented design, which grounds predictions in semantically similar, clinically validated reference cases.

\textit{\textbf{Retrieval-Augmented Inference.}}
Retrieval-augmented methods combine neural representations with external memory to support reasoning beyond end-to-end learning.
Pioneering work, such as RAG~\cite{lewis2020retrieval}, introduced a hybrid framework that retrieves documents during inference to enhance language generation.
In the medical domain, MedRAG~\cite{chen2023medrag} and similar frameworks have applied retrieval (more specifically, conditioning on relevant examples) to improve diagnosis and clinical report generation.
Related approaches in few-shot learning~\cite{vinyals2016matching, sung2018learning} and memory-based reasoning~\cite{esteva2021deep} leverage structured knowledge and prior cases to enhance generalization.
However, most of these methods do not fully integrate multimodal context or align their reasoning process with clinical workflows.
REMEMBER bridges this gap by retrieving semantically relevant, multimodal evidence and integrating it through attention-guided inference.

\begin{figure*}[]
    \centering
    \includegraphics[width=\linewidth]{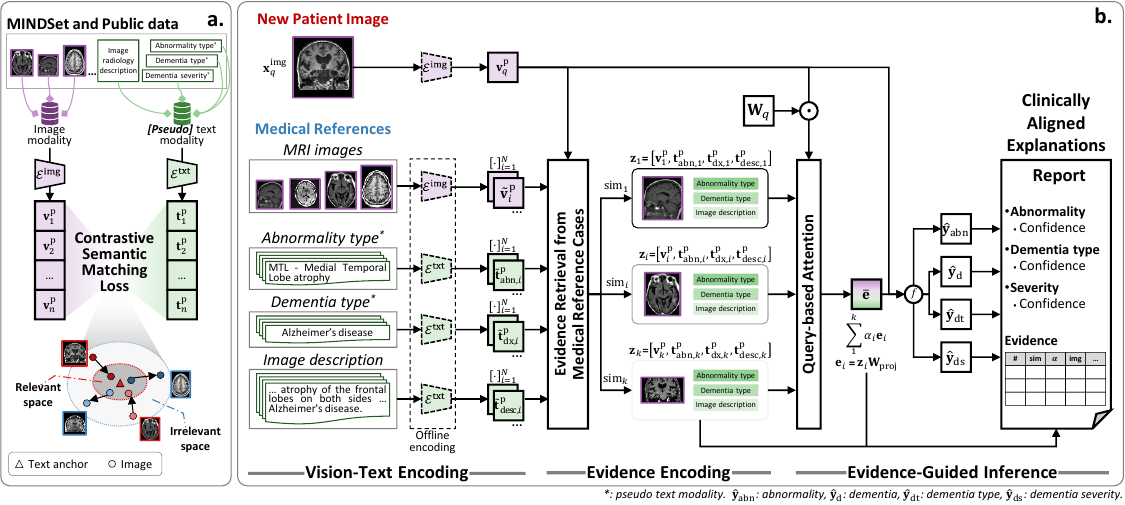}
    \caption{
        \textbf{An overview of the REMEMBER framework.}
        \normalfont
        When receiving a brain MRI scan from a new subject (a query brain scan), REMEMBER embeds the neuroimaging data and compares them to textual anchors and reference cases (confirmed cases most similar to the query brain scan).
        Subsequently, REMEMBER encodes the retrieved evidence - both imaging and textual - and uses it to make the final diagnosis via an attention-based inference module.
    }
    \Description{
        A query brain MRI slice is embedded and compared to textual anchors and reference cases (confirmed cases most similar to the query brain scan).
        REMEMBER subsequently encodes the retrieved evidence and uses it to assist the final diagnosis via an attention-based inference module.
    }
    \label{fig:framework}
\end{figure*}

\textit{\textbf{Explainability in Medical AI.}}
Explainability remains a bottleneck for adopting AI systems in clinics~\cite{tjoa2020survey,sadeghi2024review}.
% At present, there are a few techniques designed to increase the explainability of AI, including saliency maps~\cite{simonyan2013deep,wollek2023attention}, gradient-based attribution~\cite{selvaraju2017grad}, and attention visualization~\cite{jetley2018learn}.
% The explainability of these approaches, however, is restricted in that they largely peek into model behavior, yet often fail to provide human-understandable insights about the model output, such as predictions.
At present, common techniques such as saliency maps~\cite{simonyan2013deep,wollek2023attention}, gradient-based attribution~\cite{selvaraju2017grad}, and attention visualization~\cite{jetley2018learn} offer insights into model behavior, but their predictions often lack rationale understood to humans.
Recent work has explored more human-centric approaches, such as case-based reasoning~\cite{kim2014bayesian}, evidence grounding~\cite{rajpurkar2022ai}, and report generation~\cite{boag2020baselines}.
In parallel, a few models attempt to align outputs with clinical concepts, but typically do not incorporate retrieved examples -- i.e., validated cases similar to the current input -- into their reasoning pipeline.
Nevertheless, producing explanations that are both faithful to model reasoning and clinically useful remains challenging~\cite{ghassemi2020review,wysocki2023assessing}.
REMEMBER is designed to directly address this gap by grounding predictions in retrieved case-level evidence and delivering structured, clinically aligned explanations.

\textit{\textbf{Open Challenges and Research Gaps.}}
Before introducing REMEMBER, we highlight three key limitations of current diagnostic AI models.
First, most models lack \textbf{robust generalization} in zero-shot or limited-data settings. This restricts models' scalability across clinical sites, populations, and disease variants (e.g., Alzheimer's vs. vascular dementia).
Second, while vision-language models improve interpretability, they focus on aligning inputs with predefined labels rather than \textbf{reasoning over retrieved, case-level evidence}. This overlooks a core principle of clinical diagnosis -- comparing new patients with similar prior cases to support diagnosis and interpretation.
Third, \textbf{explainability remains shallow} in medical AI. Most techniques rely on visualizations (e.g., feature maps or attention weights), offering limited clinical insight. Explanations grounded in reference cases -- reflecting how clinicians reason -- are rarely explored.
To address these gaps, \textbf{REMEMBER} integrates contrastive vision-language pretraining, reference retrieval, and attention-guided inference, enabling zero-/few-shot diagnosis with transparent justifications endorsed by clinically validated examples.

% ===============================================================
% SECTION Methodology
% ===============================================================
\section{Methodology}
% PREVIOUS PLACE OF FIGURE 2

% ===============================================================
% REMEMBER Framework Overview
\subsection{An Overview of the REMEMBER Framework}
We design REMEMBER to imitate the clinical decision-making process for diagnosing neurodegenerative diseases from radiology data.
In other words, given an unseen brain MRI slice, it retrieves semantically similar reference cases (scans from confirmed cases most similar to this new image) from an expert-curated dataset and performs clinical inference (e.g., disease type, status, and severity) based on both the input image and contextualized evidence.

An overview of the architecture is in Figure~\ref{fig:framework}.
The REMEMBER pipeline consists of five stages:
(i)~\textit{Vision-text encoding}: extract latent representations of MRI slices and pseudo-clinical descriptions using a dual encoder.
(ii)~\textit{Zero-shot diagnosis}: predict abnormality, dementia type, and severity via similarity matching with predefined clinical text anchors.
(iii)~\textit{Evidence encoding}: construct a multimodal evidence matrix from embeddings of the top-$k$ retrieved image-text reference cases.
(iv)~\textit{Evidence-guided inference}: integrate the query and evidence via attention to produce final predictions.
(v)~\textit{Clinically aligned explanations}: generate structured diagnostic reports with predicted labels, confidence scores, and weighted references. 

% ===============================================================
% Vision-Text Encoder
\subsection{Vision-Text Encoder}
REMEMBER adopts a dual-encoder architecture inspired by CLIP~\cite{radford2021learning} and its biomedical variants~\cite{zhang2024biomedclip, can2025vista}.
It consists of a vision encoder $\mathcal{E}^{\text{img}}$ and a text encoder $\mathcal{E}^{\text{txt}}$.
Each encoder maps its corresponding modality into a shared latent space to facilitate image-text alignment.
Each input consists of:
A 2D axial slice $\mathbf{x}^{\text{img}} \in \mathbb{R}^{H \times W}$ extracted from a 3D brain MRI volume.
A textual description $\mathbf{x}^{\text{txt}}$ corresponding to one of several clinically relevant pseudo-modalities.

\textbf{Data construction.}
We train on combined expert-verified and public datasets:
% \textbf{MINDSet}~\cite{can2025vista}. Each sample has four image-text pairs:
% image-original radiology report,
% image-pseudo abnormality description,
% image-pseudo dementia type description,
% and image-pseudo abnormality and dementia type description.
% % 
% \textbf{Public dataset}~\cite{alzheimer_mri_dataset}. We synthesize dementia severity labels to create image-pseudo text pairs.
%
\begin{itemize}
    \item \textbf{MINDSet}~\cite{can2025vista}. Each sample has four image-text pairs:
    image-original radiology report,
    image-pseudo abnormality description,
    image-pseudo dementia type description,
    and image-pseudo abnormality and dementia type description.

    \item \textbf{Public dataset}~\cite{alzheimer_mri_dataset}. We synthesize dementia severity labels to create image-pseudo text pairs.
\end{itemize}
A description of the pseudo text modalities is in Appendix~\ref{appendix:text_modalities}.

\textbf{Contrastive objective.}
Let $\mathbf{v}_i = \mathcal{E}^{\text{img}}(\mathbf{x}^{\text{img}}_i)$ and $\mathbf{t}_i = \mathcal{E}^{\text{txt}}(\mathbf{x}^{\text{txt}}_i)$ be the embeddings of a matching image--text pair. 
Here, we optimize a contrastive loss over a batch of $N$ pairs:
\begin{equation*}
    \mathcal{L}^{\text{contrastive}} = - \frac{1}{N} \sum_{i=1}^N \log \frac{\exp\left( \text{sim}\left( \mathbf{v}_i, \mathbf{t}_i \right) / \tau \right)}{\sum_{j=1}^N \exp\left( \text{sim}\left( \mathbf{v}_i, \mathbf{t}_j \right) / \tau \right)},
\end{equation*}
where $\text{sim}(\cdot, \cdot)$ is cosine similarity and $\tau$ is a temperature parameter.

% ===============================================================
% Zero-shot Diagnosis Pipeline
\subsection{The Pipeline for Zero-shot Diagnosis}
\label{sec:zero-shot}
When making inference, REMEMBER first performs zero-shot classification by embedding the query image $\mathbf{x}_q^{\text{img}}$ into the shared latent space using a projection head $f^v$:
\begin{equation*}
    \mathbf{v}_q^{\text{p}} = f^v \left( \mathcal{E}^{\text{img}} \left( \mathbf{x}_q^{\text{img}} \right) \right).
\end{equation*}
It is then compared to pre-defined textual anchors $\left\{ \mathbf{t}_k^{\text{p}} \right\}$ corresponding to clinical labels such as:
\textit{abnormality type} (e.g., hippocampal atrophy and ventricular enlargement);
\textit{dementia diagnosis} (e.g., AD vs. non-AD);
\textit{dementia subtype} (e.g., non-demented, AD, and other dementia);
\textit{dementia severity} (e.g., very mild, mild, and moderate).

Subsequently, REMEMBER makes predictions by computing cosine similarity between $\mathbf{v}_q^{\text{p}}$ and each anchor:
\begin{equation*}
\hat{y}_{\text{zero-shot}} = \argmax_k \cos \left( \mathbf{v}_q^{\text{p}}, \mathbf{t}_k^{\text{p}} \right).
\end{equation*}
Simultaneously, REMEMBER retrieves the top-$k$ reference cases from the training medical reference MINDset using
% $\text{sim}_i = \cos \left( \mathbf{v}_q^{\text{p}}, \mathbf{v}_i^{\text{p}} \right)$.
$\text{sim}\left( \mathbf{v}_q^{\text{p}}, \mathbf{v}_i^{\text{p}} \right)$.
The mathematical formulation of REMEMBER and label mappings for each prediction task are in Appendix~\ref{appendix:zeroshot}.

% ===============================================================
% Evidence Encoding Module
\subsection{Evidence Encoding Module}

To aid few-shot diagnoses, REMEMBER uses retrieved reference cases to provide contextual cues for alignment, reasoning, and interpretability.
Particularly, the evidence encoding module integrates multimodal signals (images, abnormality labels, diagnostic descriptions, and semantic similarity) into a unified representation.

% Given a query image, we encode the top-$k$ retrieved cases as follows:

% \begin{itemize}
%     \item \textbf{Query image encoding:} The embedding $\mathbf{v}_q^{\text{p}}$ is computed using the vision encoder and projection head as described in Section~\ref{sec:zero-shot}.
    
%     \item \textbf{Reference image encoding:} For each retrieved case $i$, we encode the image using $\mathbf{v}i^{\text{p}} = f^v(\mathcal{E}^{\text{img}}(\mathbf{x}i^{\text{img}}))$,
%     and extract three textual embeddings from the abnormality-type description, dementia label description, and original radiology-style description, respectively.
%     Each text modality is processed via the shared text encoder and projection head:
%     $\mathbf{t}_{\text{abn},i}^{\text{p}}$, $\mathbf{t}_{\text{dx},i}^{\text{p}}$, and $\mathbf{t}_{\text{desc},i}^{\text{p}} = f^t(\mathcal{E}^{\text{txt}}(\text{text}))$.
% \end{itemize}

\textbf{Query image encoding:} we compute its embedding $\mathbf{v}_q^{\text{p}}$ using the vision encoder and projection head, as described in Section~\ref{sec:zero-shot}.

\textbf{Reference cases encoding:} For each of the top-$k$ retrieved cases, we encode the \textit{image} as $\mathbf{v}_i^{\text{p}} = f^v(\mathcal{E}^{\text{img}}(\mathbf{x}_i^{\text{img}}))$.
In parallel, we extract three \textit{textual embeddings} corresponding to the abnormality type, dementia label, and original radiology-style description.
These are processed through the shared text encoder and projection head, yielding $\mathbf{t}_{\text{abn},i}^{\text{p}}$, $\mathbf{t}_{\text{dx},i}^{\text{p}}$, and $\mathbf{t}_{\text{desc},i}^{\text{p}} = f^t(\mathcal{E}^{\text{txt}}(\text{text}))$, respectively.

We construct a multimodal evidence vector for each reference case $i$ by concatenating both raw and similarity-weighted features:
\begin{equation*}
    \mathbf{e}_i = \text{MLP}\left(
        \left[
        \mathbf{z}_i,\; \text{sim}_i \cdot \mathbf{z}_i
        \right]
    \right), \quad
    \text{where } \mathbf{z}_i = \left[
        \mathbf{v}_i^{\text{p}},\;
        \mathbf{t}_{\text{abn},i}^{\text{p}},\;
        \mathbf{t}_{\text{dx},i}^{\text{p}},\;
        \mathbf{t}_{\text{desc},i}^{\text{p}}
    \right].
\end{equation*}
This fusion allows REMEMBER to consider the raw evidence and its relevance to the query. REMEMBER then projects the evidence vector onto a shared latent space using a linear transformation:
\begin{equation*}
\mathbf{e}_i^{\text{proj}} = \mathbf{e}_i \cdot \mathbf{W}_{\text{proj}}.
\end{equation*}
Finally, REMEMBER stacks the encoded evidence vectors to form the full evidence matrix:
\begin{equation*}
\mathbf{E} = \left[ \mathbf{e}_1^{\text{proj}},\; \mathbf{e}_2^{\text{proj}},\; \dots,\; \mathbf{e}_k^{\text{proj}} \right] \in \mathbb{R}^{k \times D}.
\end{equation*}

% ===============================================================
% Evidence-Guided Inference Head
\subsection{Evidence-Guided Inference Head}
REMEMBER performs the final diagnosis by fusing the query representation with encoded evidence using an attention-based inference mechanism.
This allows predictions to be conditioned not only on the input image but also on clinically validated reference cases.

Given the projected query embedding $\mathbf{v}_q^{\text{p}}$, a task-specific query vector is computed via a linear transformation:
% $\mathbf{q}_v = \mathbf{v}_q^{\text{p}} \cdot \mathbf{W}_q.$
\begin{equation*}
    \mathbf{q}_v = \mathbf{v}_q^{\text{p}} \cdot \mathbf{W}_q.
\end{equation*}
REMEMBER then computes attention weights over the $k$ reference cases using dot-product attention between the query $\mathbf{q}_v$ and the evidence matrix $\mathbf{E} \in \mathbb{R}^{k \times D}$:
\begin{equation*}
    \boldsymbol{\alpha} = \text{softmax} \left( \mathbf{q}_v^\top \mathbf{E}^\top \right) \in \mathbb{R}^k.
\end{equation*}
Next, REMEMBER calculates the attention weights to compute a weighted evidence summary vector:
\begin{equation*}
    \mathbf{\bar{\mathbf{e}}} = \sum_{i=1}^k \boldsymbol{\alpha}_i \mathbf{e}_i.
\end{equation*}
Finally, REMEMBER generates the prediction by feeding the concatenation of the original query embedding and the aggregated evidence vector into a multi-layer perceptron:
\begin{equation*}
    \hat{\mathbf{y}}_{\text{few-shot}} = \text{MLP} \left(\left[ \mathbf{v}_q^{\text{p}};\; \bar{\mathbf{e}} \right]\right).
\end{equation*}

% This inference head enables REMEMBER to reason over retrieved exemplars. Specifically, by doing so, REMEMBER makes prediction by incorporating relevant past cases: this enhances REMEMBER's predictability and interpretability when making clinical decisions.
This inference head enables REMEMBER to reason over retrieved cases, incorporating relevant past cases into prediction -- enhancing both accuracy and interpretability in clinical decision-making.

% ===============================================================
% Explanation
\subsection{Clinically Aligned Explanations}

Beyond label prediction, REMEMBER generates structured diagnostic reports grounded in retrieved reference cases.
Each report includes predictions for multiple tasks (e.g., abnormality type, dementia diagnosis, and severity), along with softmax-normalized confidence scores.
To contextualize predictions, REMEMBER retrieves top-$k$ reference cases from the MINDSet corpus and estimates each reference's influence via learned attention weights.

A template of REMEMBER's report is shown in Appendix~\ref{appendix:report_template}.
Let $\hat{y}_{\text{abn}}$, $\hat{y}_{\text{type}}$, and $\hat{y}_{\text{severity}}$ denote the predicted labels for each task, with corresponding confidence vectors $\mathbf{p}_{\text{abn}}$, $\mathbf{p}_{\text{type}}$, and $\mathbf{p}_{\text{severity}}$.
For each reference $i$, let $\text{sim}_i$ be the cosine similarity, $\alpha_i$ the attention weight, $y_i^{\text{abn}}$, $y_i^{\text{dx}}$ the ground-truth labels, and $\text{text}_{\text{desc},i}$ the radiology report.
REMEMBER organizes the explanation as:
\begin{equation*}
\text{Report} = \left\{
\begin{aligned}
&\hat{y}_{\text{abn}}, \mathbf{p}_{\text{abn}},\; 
\hat{y}_{\text{type}}, \mathbf{p}_{\text{type}},\;
\hat{y}_{\text{severity}}, \mathbf{p}_{\text{severity}}, \\
&\left( \text{sim}_i,\; \alpha_i,\; y_i^{\text{abn}},\; y_i^{\text{dx}},\; \text{text}_{\text{desc},i} \right)_{i=1}^{k}
\end{aligned}
\right\}.
\end{equation*}
This format provides clinical justification by showing the similarity between the subject and confirmed cases, and quantifies each reference's contribution via attention weights. 
% By making diagnostic decisions based on real, explainable clinical examples, REMEMBER supports transparent and trustworthy reasoning that mimics practical clinical decision-making~\cite{ghassemi2020review,doshi2017towards}.
By grounding decisions in real, interpretable examples, REMEMBER supports transparent, trustworthy reasoning aligned with clinical practice~\cite{ghassemi2020review,doshi2017towards}.

% ===============================================================
% SECTION Experiments
% ===============================================================
\section{Experiments}

% ===============================================================
% Datasets
\subsection{Datasets}
\label{sec:dataset}
We evaluate REMEMBER using two datasets.
The first, MINDSet~\cite{can2025vista}, is a curated multimodal dataset; we constructed it to support abnormality identification, clinical reasoning, and retrieval-based explanation.
The second is a publicly available MRI-based Alzheimer's Disease classification dataset~\cite{alzheimer_mri_dataset}; we will use it to assess diagnostic performance in realistic scenarios.

\textbf{MINDSet Reference Dataset.}
The MINDSet dataset comprises 170 radiology brain images, each paired with structured textual descriptions and annotated abnormality types.
These cases were collected from medical literature, online radiology repositories, and textbooks, and were reviewed by human experts to ensure clinical validity.
This dataset serves as the backbone for retrieval, evidence encoding, and zero-shot generalization in REMEMBER.
% Its human-validated structure is essential for generating transparent, case-based reasoning outputs.

\textbf{Public AD Classification Dataset.}
To evaluate predictive performance, we use a public benchmark dataset containing 2D axial slices of brain MRIs labeled for dementia severity.
The dataset includes $5,120$ training and $1,280$ test samples labeled with four severity stages: \textit{non-demented}, \textit{very mild}, \textit{mild}, and \textit{moderate} dementia.
For binary dementia classification, we follow prior work and re-map the observed labels into either \textit{non-demented} or \textit{demented}. % aligning with the disease prediction objective in REMEMBER.

% Together, these two datasets provide complementary views: MINDSet enables retrieval-based reasoning and explanation, while the public dataset allows robust evaluation of diagnostic performance under zero- and few-shot conditions.

% ===============================================================
% Evaluation Setup
\subsection{Evaluation Setup}
To evaluate REMEMBER's efficacy in a relatively broad range of scenarios, we consider both \textit{zero-shot} and \textit{few-shot} tasks across four diagnostic tasks:
(i)~\textbf{abnormality type classification} (i.e., normal, MTL atrophy, WMH, and others),
(ii)~\textbf{binary dementia classification} (i.e., non-demented vs. demented),
(iii)~\textbf{dementia type classification} (i.e., non-dementia, Alzheimer's disease, and other dementia),
and (iv)~\textbf{dementia severity classification} (i.e., non-demented, very mild, mild, and moderate dementia).

\textbf{Baselines.}
We compare REMEMBER with BiomedCLIP~\cite{zhang2024biomedclip},
% MedCLIP~\cite{wang2022medclip},
ConVIRT~\cite{zhang2022contrastive}, and VisTA~\cite{can2025vista}, state-of-the-art vision-language models in this domain.
We evaluate the models under the same zero-shot protocol using cosine similarity to text anchors for classification.

\textbf{Metrics.}
To quantify the classification performance, we report a broad range of metrics: Accuracy, Precision, Recall (Sensitivity), F1 Score, and Specificity. %, and Area Under the ROC Curve (AUC-ROC).
Unless otherwise specified, we macro-averaged all metrics across classes.

\textbf{Computational Setup.}
We conduct all experiments on a single NVIDIA TESLA P100 GPU using Kaggle's free GPU runtime.
We design REMEMBER to be efficient, such that it does not require large-scale training beyond vision-text pretraining.
For example, the inference time per image is under $200$ms.
Hyperparameter details and training configurations are in Appendix~\ref{appendix:hyperparameters}.

\textbf{Code and Model Availability.}  
To ensure transparency and reproducibility, we make our implementation publicly available at \href{https://anonymous.4open.science/r/remember}{Anonymous GitHub repository}.  
We will also release the pretrained REMEMBER model at the \href{https://huggingface.co}{Hugging Face platform}.

% =================================================================
% Overall Performance
\subsection{Model Performance}
\label{sec:overall-performance}

\begin{table}[]
\centering
\caption{
    \textbf{REMEMBER's performance across diagnostic tasks.}
    \normalfont
    We report the performances of three baseline vision-language models (BiomedCLIP, ConVIRT, VisTA) and our proposed REMEMBER model under two configurations: zero-shot diagnosis and evidence-guided inference.
    Results are shown for both the MINDSet and public AD dataset, depending on task.
    REMEMBER consistently outperforms baselines in both settings and across tasks. %, demonstrating strong generalization, evidence integration, and interpretability.
}
\label{tab:overall_result}
\resizebox{\linewidth}{!}{%
\begin{tabular}{lccccc}
\hline
\multicolumn{1}{c}{\textbf{Methods}} & \textbf{\begin{tabular}[c]{@{}c@{}}\texttt{\;ACC\;}\\ (\%)\end{tabular}} & \textbf{\begin{tabular}[c]{@{}c@{}}\texttt{\;\;P\;\;}\\ (\%)\end{tabular}} & \textbf{\begin{tabular}[c]{@{}c@{}}\texttt{\;\;R\;\;}\\ (\%)\end{tabular}} & \textbf{\begin{tabular}[c]{@{}c@{}}\texttt{\;\;F1\;\;}\\ (\%)\end{tabular}} & \textbf{\begin{tabular}[c]{@{}c@{}}\texttt{SPEC}\\ (\%)\end{tabular}} \\ \hline
\multicolumn{6}{l}{\cellcolor[HTML]{C0C0C0}\textbf{a. Abnormality type prediction}$^\dagger$} \\ \hline
\multicolumn{6}{l}{{\color[HTML]{329A9D} \textit{Performance on MINDSet}}} \\ \hline
BiomedCLIP~\cite{zhang2024biomedclip} & 26.00 & 25.96 & 37.85 & 27.39 & 80.04 \\
% MedCLIP~\cite{wang2022medclip} & 24.00 & 36.14 & 22.96 & 18.23 & 75.51 \\
ConVIRT~\cite{zhang2022contrastive} & 26.00 & 6.50 & 25.00 & 10.32 & 75.00 \\
VisTA~\cite{can2025vista} & 74.00 & 74.02 & 72.30 & 72.05 & 90.53 \\
\multicolumn{6}{l}{\textbf{REMEMBER}$^\S$ \textit{(our proposed model)}} \\
\quad-- zero-shot & 84.00 & 81.47 & \textbf{92.00} & 86.42& 95.16 \\
\quad-- evidence-guided & \textbf{88.00} & \textbf{86.29} & 90.18 & \textbf{88.19}& \textbf{95.37} \\ \hline
\multicolumn{6}{l}{\cellcolor[HTML]{C0C0C0}\textbf{b. Binary dementia classification}$^\ddagger$} \\ \hline
\multicolumn{6}{l}{{\color[HTML]{329A9D} \textit{Performance on MINDSet}}} \\ \hline
BiomedCLIP~\cite{zhang2024biomedclip} & 30.00 & 77.77 & 17.50 & 28.57 & 80.00 \\
% MedCLIP~\cite{wang2022medclip} & 58.00 & 78.78 & 65.00 & 71.23 & 30.00 \\
ConVIRT~\cite{zhang2022contrastive} & 80.00 & 80.00 & \textbf{100.00} & 88.88 & 0.00 \\
VisTA~\cite{can2025vista} & 88.00 & 90.48 & 95.00 & 92.68 & 60.00 \\
\multicolumn{6}{l}{\textbf{REMEMBER}$^\S$ \textit{(our proposed model)}} \\
\quad-- zero-shot & 94.00 & 95.12 & 97.50 & 96.30 & 80.00 \\
\quad-- evidence-guided & \textbf{96.00} & \textbf{97.50} & 97.50 & \textbf{97.50} & \textbf{90.00} \\ \hline
\multicolumn{6}{l}{{\color[HTML]{F8A102} \textit{Performance on public dataset}}} \\ \hline
BiomedCLIP~\cite{zhang2024biomedclip} & 49.45 & \textbf{100.00} & 0.15 & 0.31 & \textbf{99.68} \\
% MedCLIP~\cite{wang2022medclip} & 50.70 & 51.36 & 43.81 & 47.28 & 57.73 \\
ConVIRT~\cite{zhang2022contrastive} & 50.47 & 50.46 & \textbf{100.00} & 67.08 & 0.00 \\
VisTA~\cite{can2025vista} & 51.64 & 51.11 & \textbf{100.00} & 67.64 & 2.40 \\
\multicolumn{6}{l}{\textbf{REMEMBER}$^\S$ \textit{(our proposed model)}} \\
\quad-- zero-shot & 97.19 & 98.57 & 95.82 & 97.18& 98.58 \\
\quad-- evidence-guided & \textbf{97.27} & 97.66 & 96.90 & \textbf{97.28} & 97.63 \\ \hline
\multicolumn{6}{l}{\cellcolor[HTML]{C0C0C0}\textbf{c. Dementia type classification}$^\dagger$} \\ \hline
\multicolumn{6}{l}{{\color[HTML]{329A9D} \textit{Performance on MINDSet}}} \\ \hline
BiomedCLIP~\cite{zhang2024biomedclip} & 30.00 & 32.96 & 42.09 & 36.97& 68.33 \\
% MedCLIP~\cite{wang2022medclip} &  &  &  &  &  \\
ConVIRT~\cite{zhang2022contrastive} & 34.00 & 60.14 & 39.23 & 47.48 & 69.54 \\
VisTA~\cite{can2025vista} & 42.00 & 46.56 & 51.03 & 48.69& 71.39 \\
\multicolumn{6}{l}{\textbf{REMEMBER}$^\S$ \textit{(our proposed model)}} \\
\quad-- zero-shot & 42.00 & 48.56 & 55.42 & 51.76& 71.48 \\
\quad-- evidence-guided & \textbf{86.00} & \textbf{87.37} & \textbf{86.63} & \textbf{87.00}& \textbf{91.67} \\ \hline
\multicolumn{6}{l}{\cellcolor[HTML]{C0C0C0}\textbf{d. Dementia severity classification}$^\dagger$} \\ \hline
\multicolumn{6}{l}{{\color[HTML]{F8A102} \textit{Performance on public dataset}}} \\ \hline
BiomedCLIP~\cite{zhang2024biomedclip} & 47.58 & 31.60 & 25.29 & 28.10& 75.34 \\
% MedCLIP~\cite{wang2022medclip} &  &  &  &  &  \\
ConVIRT~\cite{zhang2022contrastive} & 36.17 & 29.78 & 25.28 & 27.35 & 75.09 \\
VisTA~\cite{can2025vista} & 38.44 & 33.09 & 30.01 & 31.47& 77.71 \\
\multicolumn{6}{l}{\textbf{REMEMBER}$^\S$ \textit{(our proposed model)}} \\
\quad-- zero-shot & 90.94 & \textbf{73.04} & \textbf{72.63} & \textbf{72.83}& 96.89 \\
\quad-- evidence-guided\quad\quad\;\; & \textbf{94.06} & 69.18 & 69.72 & 69.45& \textbf{97.69} \\ \hline
\multicolumn{6}{r}{\footnotesize\begin{tabular}[c]{@{}r@{}}
    \texttt{ACC}: accuracy, \texttt{P}: precision, \texttt{R}: recall (sensitivity), \texttt{F1}: F1 score, \texttt{SPEC}: specificity.\\
    The best results in each column and experiment are in bold fonts.\\
    $^\dagger$: Macro-averaged results across classes. 
    $^\ddagger$: Binary classification results.\\
    $^\S$: REMEMBER models using zero-shot diagnosis pipeline and evidence-guided inference head.
\end{tabular}}
\end{tabular}%
}
\end{table}

We compare REMEMBER with present state-of-the-art methods across four diagnostic tasks: abnormality type prediction, binary dementia classification, dementia type classification, and dementia severity classification.
The results are in Table~\ref{tab:overall_result}.

\textbf{Abnormality Type Prediction} ($4$-class prediction).  
On the MINDSet dataset, REMEMBER significantly outperforms all evaluated models in zero-shot and evidence-guided inference.
For zero-shot tasks, REMEMBER achieves accuracy, F1 score, and specificity of $84\%$, $83.79\%$, and $95.16\%$, respectively. It exceeds VisTA by $14\%$ in F1.
When a small labeled set is available, accuracy rises to $88\%$.
Conversely, models such as BiomedCLIP and ConVIRT achieve chance-level accuracies;
this hints their potential limitation in handling detailed abnormality reasoning without task-specific tuning.

\textbf{Dementia Classification} (binary prediction).
Our model achieves robust performance on both the MINDSet and public datasets.
On MINDSet data, REMEMBER reaches $94\%$ accuracy and $96.3\%$ F1 score under the zero-shot setting, outperforming all evaluated methods.
REMEMBER further improves F1 to $97.5\%$ with evidence-guided inference.
On the public dataset, REMEMBER achieves an accuracy of  $97.19\%$ in the zero-shot setting, significantly outperforming VisTA ($51.64\%$) and BiomedCLIP ($49.45\%$).
In general, other models struggle with generalization: they either show high precision but extremely low recall, and vice versa.
REMEMBER, in comparison, balances both sensitivity ($95.82\%$) and specificity ($98.58\%$).

\textbf{Dementia Type Classification} ($3$-class prediction).
This task consider a three-class classification.
REMEMBER demonstrates strong performance, particularly with evidence-guided inference.
It achieves 86\% accuracy and $86.96\%$ F1 score, a $40\%$ improvement over VisTA.
While all baseline methods struggle in the zero-shot setting, REMEMBER maintains stability even before adaptation.

\textbf{Dementia Severity Classification} ($4$-class prediction).
This task, performed on the public dataset, is more fine-grained with four severity levels.
In the zero-shot setting, REMEMBER achieves $90.94\%$ accuracy and $69.63\%$ F1 score.
The second best method, VisTA, only achieves $38.44\%$ accuracy and $28.38\%$ F1.
With evidence-guided inference, REMEMBER further improves and obtains 94.06\% accuracy.
These results suggest REMEMBER's ability to distinguish subtle distinctions between groups with finely defined differences: non-demented, very mild, mild, and moderate dementia. 

% \textbf{Summary.}  
% Across all tasks and settings, REMEMBER consistently outperforms existing baselines, particularly in low-resource scenarios.
% Its ability to leverage vision-text alignment and retrieved evidence allows it to maintain high diagnostic accuracy while producing explainable outputs.
% This affirms its utility in real-world clinical contexts where annotated data is often limited.

% ===============================================================
% Few-shot
\subsection{Few-Shot Generalization}
\label{sec:few-shot}

\begin{figure}[]
    \centering
    \includegraphics[width=\linewidth]{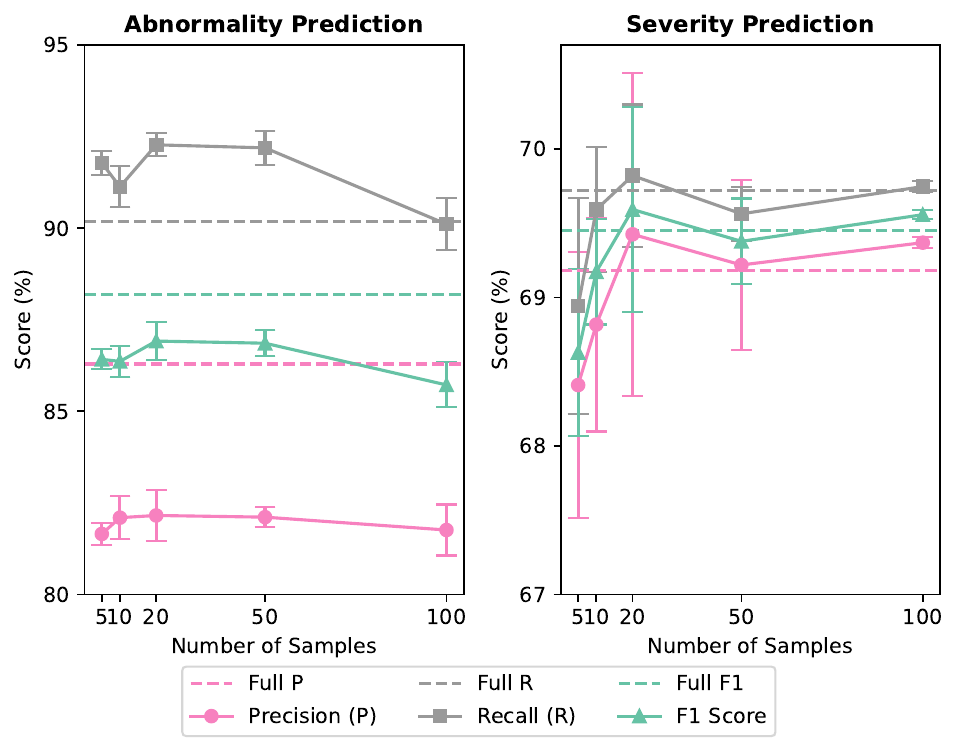}
    \caption{
        \textbf{Performance of REMEMBER under few-shot supervision for two representative tasks.}
        \normalfont
        Evaluation of REMEMBER with varying numbers of labeled training samples per class ($k \in \{5, 10, 20, 50, 100\}$).
        Left: Abnormality prediction on the curated MINDSet dataset. 
        Right: Dementia severity prediction on the public dataset. 
        The mean and standard deviation of Precision, Recall, and F1 are reported across 10 runs. 
        REMEMBER achieves robust performance and stable generalization even with limited supervision.
    }
    \Description{
        Evaluation of REMEMBER with varying numbers of labeled training samples per class ($k \in \{5, 10, 20, 50, 100\}$).
        Left: abnormality prediction on MINDSet. 
        Right: dementia severity prediction on the public dataset. 
        Mean and standard deviation of Precision, Recall, and F1 are reported across 10 runs. 
        REMEMBER maintains competitive performance and stable generalization even with limited supervision.
    }
    \label{fig:fewshot_results}
\end{figure}

To evaluate REMEMBER's performance under low-resource conditions, we conduct a controlled few-shot study by varying the number of labeled training examples per class.
We experiment with $k = \{5, 10, 20, 50, 100\}$ samples per class and report the mean and standard deviation of macro-averaged metrics over 10 independent runs.
Results for two representative tasks -- abnormality type prediction (on MINDSet) and dementia severity classification (on the public dataset) -- are shown in Figure~\ref{fig:fewshot_results}.

\textbf{Abnormality prediction (MINDSet).}  
Despite MINDSet's small training size ($n = 150$), REMEMBER achieves strong performance even with as few as 5 samples/class.
At $k=20$, the model reaches $86.36\%$ F1, closely matching the full-training F1 of $88.19\%$.
Increasing $k$ to 20 or 50 further stabilizes performance, with standard deviations consistently under $2\%$.
Interestingly, larger $k$ values (e.g., 100) do not yield meaningful gains -- possibly due to data imbalance, where some minor classes have fewer than 20 training examples.

\textbf{Dementia severity prediction (Public dataset).}  
REMEMBER also generalizes well in few-shot settings on the larger public dataset.
With only 10 samples per class, the model achieves $69.17\%$ F1 -- comparable to $69.45\%$ using the full-training set.
As $k$ increases, performance remains consistent, with minimal variance at $k=100$ (F1 stdev $<0.1$), suggesting high reliability and robust convergence.

\textbf{Confidence and stability.}  
Standard deviation across runs decreases steadily with increasing $k$, indicating improved confidence and reduced sensitivity to random data splits.
This stability reinforces REMEMBER's practical applicability in real-world medical scenarios where labeled data is limited or expensive to obtain.

% ===============================================================
% Evidence Quality and Similarity Consistency
\subsection{Evidence Quality and Similarity Consistency}
\label{sec:retrieval_analysis}

\begin{figure}[]
    \centering
    \includegraphics[width=\linewidth]{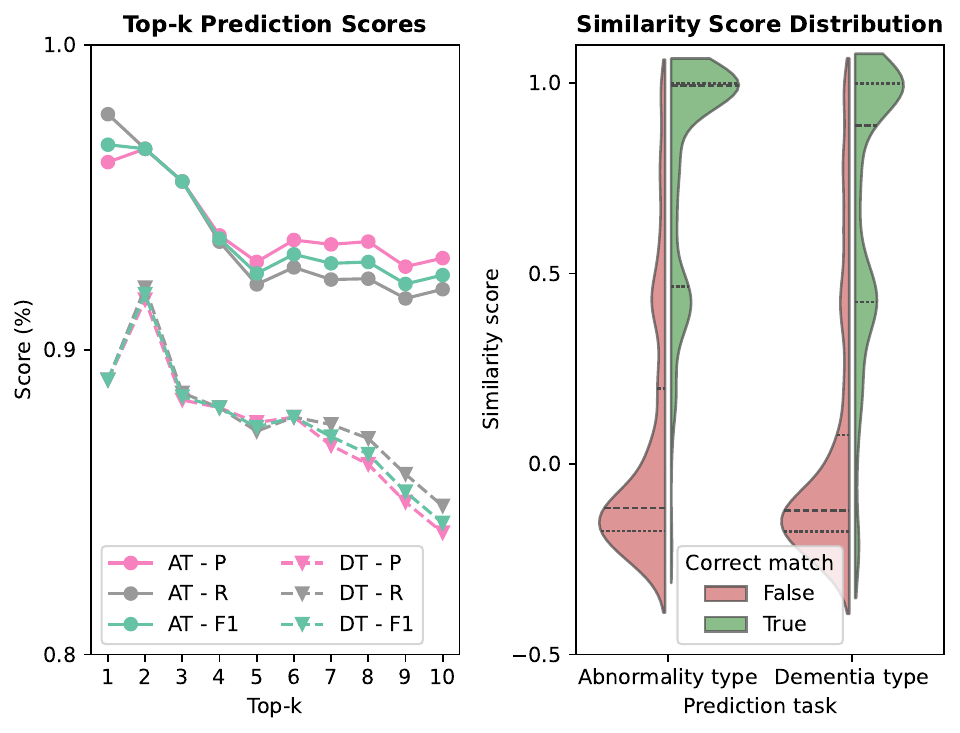}
    \caption{
        \textbf{Retrieval consistency and similarity distribution.}
        \normalfont
        Left: Label consistency of retrieved references across top-$k$ results.
        Precision, recall, and F1 scores measure how often the brain scans from the retrieved cases share the same abnormality type (AT) or dementia type (DT) as the query brain scan. 
        Right: Distribution of similarity scores. Violin plots compare the similarity scores of retrieved cases with correct vs. incorrect labels in abnormality and dementia tasks.
    }
    \Description{
        (Left) Label consistency of retrieved references across top-$k$ results.
        Precision, recall, and F1 scores measure how often the retrieved cases share the same abnormality or dementia type as the query. 
        (Right) Distribution of similarity scores. Violin plots compare the similarity scores of retrieved cases with correct vs. incorrect labels in abnormality and dementia tasks.
    }
    \label{fig:similarity_analysis}
\end{figure}

After demonstrating REMEMBER's capacity in predicting disease outcomes, we assess the semantic correctness of REMEMBER's retrieved evidence.
To that end, we evaluate whether the top-$k$ reference cases share the same abnormality type and dementia type as the query input.
We report precision, recall, and F1 score to quantify the consistency of the evidence-label for $k = 1$ to $10$.
We summarize the results in the left panel of Figure~\ref{fig:similarity_analysis}.

\textbf{Abnormality label alignment.}
REMEMBER shows high fidelity in retrieved references for abnormality type matching.
At $k=1$, it obtains precision of $96.2\%$ and recall $97.7\%$, with a peak F1 of $96.7\%$.
Even as $k$ goes to $10$, F1 remains above $92\%$, suggesting that the retrieval consistently selects structurally similar cases.
This confirms the contrastive vision-text encoder's capacity to meaningfully anchor image features with clinically relevant abnormalities.

\textbf{Dementia label alignment.}
REMEMBER also shows strong performance in aligning dementia-type, although slightly lower than structural abnormality matching.
Top-$1$ retrieved cases match the query label with an F1 of $89.0\%$.
Performance peaks at $k=2$ (F1 = $91.8\%$), then gradually declines, indicating reduced label consistency when REMEMBER considers less similar dementia cases in the retrieval list.
Nonetheless, even at $k=10$, F1 remains above $84\%$. This reinforces REMEMBER's semantic reasoning capacity.

\textbf{Similarity score distributions.}
To further understand the latent embedding space REMEMBER has learned, we visualize the distribution of similarity scores between the query image and retrieved image references, and stratify them according to whether the retrieved case shares the correct label.
The right panel of Figure~\ref{fig:similarity_analysis} presents violin plots for both tasks (abnormality and dementia type).
In both cases, correctly matched references show significantly higher cosine similarity scores than mismatched ones. This confirms that the similarity metric between query image and retrieved images correlates strongly with semantic correctness.

% ===============================================================
% Embedding Space
\subsection{Studying REMEMBER's Embedding Space}

\begin{figure}[t]
    \centering
    \includegraphics[width=\linewidth]{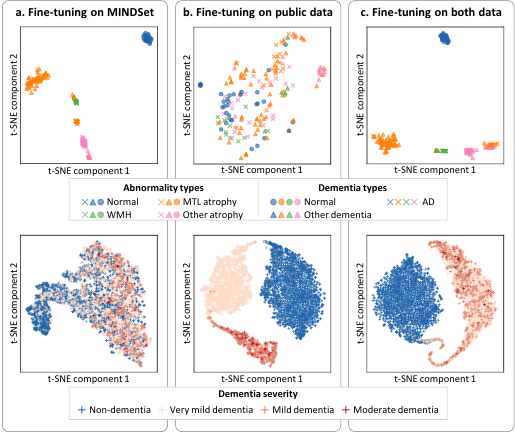}
    \caption{
        \textbf{A visualization of REMEMBER's embedding space under different fine-tuning scenarios using t-SNE.}
        \normalfont
        Each column shows the model trained on (a) the MINDSet dataset, (b) the public dataset, and (c) both datasets.
        The top row visualizes embeddings of MINDSet samples, colored by abnormality type and shaped by dementia type.
        The bottom row shows embeddings of the samples from the public dataset, colored by dementia severity.
        Models trained on a single dataset capture dataset-specific structure:
        (a) clearly separates abnormality and dementia types; (b) forms distinct clusters across dementia severity.
        Joint training (c) preserves strong separation for abnormality and dementia types while partially capturing severity as a smooth progression from non-demented to moderately demented stages.
    }
    \Description{
        Each column shows the model trained on (a) MINDSet only, (b) public data only, and (c) both datasets.
        The top row visualizes embeddings of MINDSet samples, colored by abnormality type and shaped by dementia type.
        The bottom row shows embeddings of public dataset samples, colored by dementia severity.
        Models trained on a single dataset capture dataset-specific structure:
        (a) clearly separates abnormality and dementia types; (b) forms distinct clusters across dementia severity.
        Joint training (c) preserves strong separation for abnormality and dementia types while partially capturing severity as a smooth progression from non-demented to moderate stages.
    }
    \label{fig:embedding_tsne}
\end{figure}

We assess how training data impact the representation space's structure by visualizing the extracted embeddings using t-SNE~\cite{van2008visualizing}. 
Specifically, Figure~\ref{fig:embedding_tsne} compares REMEMBER fine-tuned on: (a) MINDSet only, (b) public data only, and (c) both datasets.

In the MINDSet data (top row), fine-tuning only on MINDSet results in well-separated clusters by abnormality type (indicated by colors), with some alignment to dementia type (indicated by markers).
This suggests REMEMBER's strong capacity to capture structural imaging patterns specific to diagnostic categories.
However, this model fails to generalize well to classify dementia severity levels, as seen in the lower left panel.

In contrast, training only on the public dataset improves the separation of severity levels (bottom center), showing well-separated clusters across dementia severity levels.
Yet, this model exhibits weak clustering performance for abnormality types and dementia types (top center), indicating limited cross-type generalization. % \textcolor{red}{What do you mean by cross-type?} 

When fine-tuned on both datasets (right panels), REMEMBER balances both objectives. Specifically, REMEMBER clearly clustered abnormality types and dementia categories (top right); particularly, it results in a clear separation between non-demented and demented groups for dementia severity (bottom right).
Although the intra-dementia severity transition (from very mild to moderate) is less smooth compared to the model fine-tuned using only the public data, it seems to preserve some of the within-disease trend.

Together, these results show REMEMBER learns a representation space that generalizes across data sources and preserves clinically meaningful structure for classification and disease staging.

% ===============================================================
% Ablation
\subsection{Ablation Study}
\label{sec:ablation}

\begin{figure}[]
    \centering
    \includegraphics[width=\linewidth]{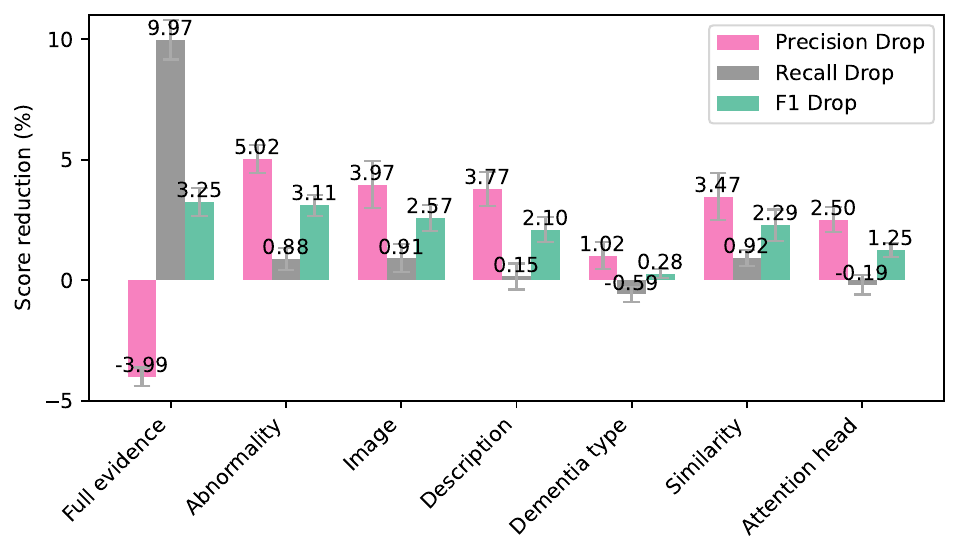}
    \caption{
        \textbf{Ablation study on abnormality prediction.}
        \normalfont
         When removing components from REMEMBER's evidence-guided inference pipeline (where the components include the evidence image, textual modalities, similarity scores, and attention mechanism), model performance drops in macro-averaged F1, Precision, and Recall.
        Error bars show standard deviation across 10 runs.
    }
    \Description{
        Performance drop in macro-averaged F1, Precision, and Recall after removing components from REMEMBER's evidence-guided inference pipeline.
        Error bars show standard deviation across 10 runs.
    }
    \label{fig:ablation}
\end{figure}

To evaluate the individual contribution of the evidence image, textual modalities, similarity scores, and attention mechanism in the REMEMBER pipeline, we conduct an ablation study on the abnormality prediction task.
For each variant, we remove a specific input feature or architectural component and measure the performance drop in macro-averaged Precision, Recall, and F1 score.
We average the results over $10$ random seeds and present them in Figure~\ref{fig:ablation}.

\textbf{Full evidence vector.}  
Removing the entire evidence vector and relying solely on the query image results in the largest drop in overall performance (-$3.25\%$ F1).
Interestingly, while the evidence-augmented model achieves significantly higher recall (+$9.97\%$), it slightly reduces precision (-$3.99\%$).
This suggests that the inclusion of retrieved reference cases helps the model to identify a broader range of abnormal cases -- thereby improving sensitivity.
Such trade-offs are often desirable in clinical settings, where recall (sensitivity) is critical to minimize false negatives~\cite{rajpurkar2022ai}.

\textbf{Feature modalities.}  
Among the four components of the evidence vector, the abnormality-type, image embedding, and original radiology-style description contribute the most to abnormality classification.
Ablating the abnormality feature alone results in a $3.11\%$ F1 drop, while removing the image and description features leads to drops of $2.57\%$ and $2.10\%$, respectively.
In comparison, the dementia-type label has a minimal impact on this task.

\textbf{Similarity weighting.}  
Excluding similarity scores from the evidence vector also degrades performance (-$2.29\%$ F1). This shows that retrieval score, when integrated as a feature, is useful for selecting suitable references and beneficial for downstream inference.

\textbf{Attention mechanism.}  
Disabling the attention head and replacing it with a simple concatenation of the evidence vectors reduces performance by $1.25\%$ in F1. This implies the effective role of attention in weighting relevant reference cases.

\textbf{Generalization to other tasks.}  
We observe similar trends in the dementia-type and binary dementia classification tasks, where all evidence components contribute to improved performance.
However, for dementia severity prediction, the incorporation of reference evidence does not improve significantly over the embedding of the raw query.
We hypothesize that this is likely due to a lack of fine-grained staging information related to severity progression in the evidence pool.
This points to a promising future direction: expanding the reference set with longitudinal imaging and/or including clinical notes recorded over time for the same subject to improve fine-grained disease modeling.

\textbf{Varying the number of retrieved cases.}  
We evaluate the effect of varying the number of retrieved reference cases ($k$) on model performance.
We find that increasing $k$ beyond 3 does not lead to further improvement: the overall performance remains stable.
This suggests that the attention head can effectively down-weight irrelevant references and focusing on the most informative ones.

\vspace{-1em}
% ===============================================================
% Error analysis
\subsection{Error Analysis}
Here, we investigate the limitations of REMEMBER.
Specifically, we conduct a detailed error analysis on the test set across all four diagnostic tasks.
We examine failure cases where the predicted label differs from the ground truth and analyze patterns based on prediction confidence, attention weights, and reference descriptions.

\textbf{Low-confidence misclassifications.}
Many incorrect predictions occur when the model assigns similar confidence scores to multiple competing classes (e.g., AD vs. Other Dementia), reflecting diagnostic ambiguity.
This uncertainty is also evident in the retrieved references: attention is divided among cases with overlapping structural features (e.g., co-occurring WMH and MTL atrophy).

\textbf{Reference disagreement.}
In some cases, the highest-attention reference cases have inconsistent labels to the true diagnosis.
This may reflect the underlying clinical complexity (e.g., mild cases exhibiting both atrophic and vascular patterns) or a lack of diversity in the training references.

\textbf{Interpretability despite error.}
Even for the cases where REMEMBER made incorrect predictions, the generated explanation for these cases surprisingly often includes references that highlight relevant abnormalities, such as hippocampal shrinkage or ventricular enlargement.
This suggests that REMEMBER may still unveil clinically useful insights, even in borderline or ambiguous cases.

\textbf{Task-specific patterns.}
We find that severity classification is particularly sensitive to subtle anatomical differences that may be hard to distinguish in 2D slices.
In contrast, binary dementia classification is more robust, with most errors occurring near the non-demented / very mild decision boundary.

These findings motivate directions/tasks for future work. For example, one can work on multi-slice or volumetric modeling, expand the reference dataset, and incorporate richer clinical metadata (e.g., age, gender, ethnic information, medical history, and cognitive scores) to further disambiguate borderline cases.

% ================================================================
% SECTION Discussion
% ================================================================
\section{Discussion}
Our analyses suggest several practical advantages that REMEMBER brings to clinical AI. They also point out a few limitations and areas for future development.
Below, we summarize REMEMBER's core strengths, limitations, and potential future extensions.

Key \textbf{strengths} of REMEMBER:
\textbf{Generalizability:} REMEMBER performs well across multiple diagnostic tasks in both zero- and few-shot settings. It demonstrates robustness without relying on extensive supervision.
\textbf{Clinical relevance:} The retrieval-based inference pipeline mimics how clinicians reason in practice -- by comparing new cases to existing annotated examples. This makes REMEMBER model naturally aligned with in-clinic diagnostic workflows~\cite{ghassemi2020review}.
\textbf{Transparency and explainability:} REMEMBER produces structured, reference-backed reports with interpretable attention scores. This enables users to understand and trace model decisions~\cite{doshi2017towards}.

\textbf{Limitations} of REMEMBER:
\textbf{Retrieval quality dependency:} REMEMBER's accuracy and interpretability are sensitive to reference dataset's diversity (e.g., coverage of disease types). Poorly matched or underrepresented cases may lead to suboptimal predictions.
\textbf{Scalability:} The current retrieval mechanism operates on all stored embeddings in memory. This may limit performance when studying massive datasets or making multimodal references~\cite{izacard2022few}.

\textbf{Future directions}:
\textbf{Longitudinal modeling:} Extending REMEMBER to capture temporal disease progression using MRI data collected over time could improve disease staging and personalized monitoring~\cite{aberathne2023detection}.
\textbf{3D volumetric modeling:} While REMEMBER currently operates on 2D slices, leveraging full 3D brain volumes may enhance the detection of spatial patterns that are subtle or discontinuous in 2D. This may improve diagnostic precision.
\textbf{Multimodal expansion:} Integrating multimodal data, such as genetics, cognitive assessments, or EHRs, encoding complementary disease variability, may enhance diagnostic accuracy and endorse applications for precision medicine~\cite{tang2024multimodal}.
\textbf{Severity-aware evidence curation:} Current evidence retrieval may lack granularity in staging. Future work can include explicitly curating and incorporating severity-aligned reference cases to better support dementia progression modeling.
\textbf{Improved retrieval strategies:} Incorporating adaptive or supervised retrieval mechanisms may help refine the selection of reference cases, particularly in clinically ambiguous scenarios.
\textbf{Hierarchical or ranking loss objectives:} To better reflect clinical hierarchies (e.g., the levels of dementia severity), we plan to experiment with hierarchical or ranking-based losses that encourage embeddings in the latent space to preserve ordinal relationships.

% ================================================================
% SECTION CONCLUSION
% ================================================================
\section{Conclusion}
In this paper, we introduce \textbf{REMEMBER}, a retrieval-based, explainable machine learning framework for zero- and few-shot Alzheimer's disease diagnosis using neuroimaging data. 
REMEMBER mimics the clinical reasoning process of human experts by integrating visual features from new patient data with semantically relevant, confirmed reference cases.
Specifically, REMEMBER combines a contrastively trained vision-text encoder with attention-guided inference mechanism; this promotes making interpretable predictions without relying on large-scale labeled datasets commonly required by conventional clinical AI methods.

Beyond achieving high diagnostic performance, REMEMBER generates structured, reference-backed text reports that provide clinically meaningful explanations.
This enhances transparency and clinical adoption.
This characteristic allows REMEMBER to make potential contributions to building trustworthy, evidence-aware AI systems in healthcare.

Finally, REMEMBER offers a flexible architectural foundation for multimodal clinical reasoning where data scarcity and explainability are critical and resources are limited.
Its modular design supports future extensions, such as longitudinal disease modeling, 3D volumetric analysis, and integration with other modalities such as genetics or EHRs.
We hope that our endeavors here brings some insights into our broad field's ever-growing effort to design generalizable medical AI with human-like clinical reasoning ability.

% We hope this work contributes to ongoing efforts to design generalizable, evidence-aware medical AI systems that reason more like human clinicians—transparent, grounded, and adaptive.

\begin{acks}
We thank Kaggle for providing free GPU resources that enabled the training and evaluation of our models.
\end{acks}

\bibliographystyle{ACM-Reference-Format}
\bibliography{refs}

\ifshowappendix
% \clearpage
\appendix
% =================================================================
% APPENDIX Pseudo Text Modalities
% =================================================================
\section{Pseudo Text Modalities}
\label{appendix:text_modalities}

To enhance vision-language alignment and improve zero-/few-shot generalization, REMEMBER is trained on a diverse set of pseudo text modalities that simulate clinically relevant reporting language. These modalities are derived from both expert-verified and publicly available datasets, as described below.

\subsection{MINDSet Dataset}
\label{appendix:data_M}
Each MRI image in MINDSet is paired with four types of pseudo text descriptions, yielding four distinct training samples per image:

\subsubsection{M.A. Description Text.}
Free-form radiology-style report for the image, written or verified by medical experts.

\subsubsection{M.B. Abnormality-Type Descriptions (4 classes):}
\label{appendix:data_MB}
\begin{itemize}
    \item \textbf{Normal:} ``MRI image shows normal brain without evidence of significant structures or pathological changes.''
    \item \textbf{MTL Atrophy:} ``MRI image illustrates volume reduction and structural atrophy in the medial temporal lobes, including hippocampal shrinkage.''
    \item \textbf{WMH:} ``MRI image reveals hyperintense lesions within cerebral white matter regions, indicating white matter hyperintensities.''
    \item \textbf{Other Atrophy:} ``MRI image indicates brain atrophy in cortical or subcortical regions other than medial temporal lobes, with notable structural volume loss.''
\end{itemize}

\subsubsection{M.C. Dementia Label Descriptions (3 classes):}
\label{appendix:data_MC}
\begin{itemize}
    \item \textbf{Non-dementia:} ``MRI image presents no evident dementia-related structural changes, reflecting a normal cognitive state.''
    \item \textbf{Alzheimer's Disease (AD):} ``MRI image shows characteristic patterns of brain atrophy suggestive of Alzheimer's Disease pathology.''
    \item \textbf{Other Dementia:} ``MRI image shows structural brain abnormalities indicative of dementia types other than Alzheimer's Disease, such as Vascular dementia or Dementia with Lewy bodies.''
\end{itemize}

\subsubsection{M.D. Combined Description (B + C)}
\label{appendix:data_MD}
Concatenation of the \texttt{M.B.} abnormality description (\ref{appendix:data_MB}) and \texttt{M.C.} dementia label description (\ref{appendix:data_MC}), providing richer clinical context.

\subsection{P. Public Dataset}
\label{appendix:data_P}
Each image in the public dataset is paired with one pseudo-modality based on dementia severity classification. The corresponding descriptions are as follows:

\begin{itemize}
    \item \textbf{Non-Demented:} ``MRI image depicts normal brain anatomy without visible dementia-related atrophic or pathological changes.''
    \item \textbf{Very Mild Demented:} ``MRI image presents subtle and minimal structural changes, consistent with very mild cognitive impairment or early-stage dementia.''
    \item \textbf{Mild Demented:} ``MRI image illustrates noticeable atrophic changes in brain regions, indicative of mild dementia progression.''
    \item \textbf{Moderate Demented:} ``MRI image shows pronounced structural atrophy and pathological changes characteristic of moderate dementia severity.''
\end{itemize}

These pseudo-text modalities are designed to reflect the diversity of descriptions encountered in clinical practice, enabling the model to align image features with semantically rich textual representations.

% =================================================================
% APPENDIX Detailed Explanation of Zero-shot Diagnosis Pipeline
% =================================================================
\section{Detailed Explanation of Zero-shot Diagnosis Pipeline}
\label{appendix:zeroshot}

The REMEMBER framework performs four zero-shot classification tasks using a common query embedding $\mathbf{v}_q^{\text{p}} \in \mathbb{R}^D$ extracted from a 2D axial MRI slice.
Each task compares the query embedding to a different set of text anchor embeddings $\left\{ \mathbf{t}_k^{\text{p}} \right\}$, pre-encoded from medically validated descriptions.

\subsection{Abnormality Type Prediction}
To identify structural abnormalities, REMEMBER classifies each query image into one of four predefined abnormality categories based on the pseudo text modality \texttt{M.B} (see Appendix~\ref{appendix:data_MB}):
\begin{itemize}
    \item Normal brain
    \item Medial Temporal Lobe (MTL) Atrophy
    \item White Matter Hyperintensities (WMH)
    \item Other Atrophy
\end{itemize}

Given the projected query image embedding $\mathbf{v}_q^{\text{p}}$ and the abnormality anchor embeddings $\left\{ \mathbf{t}_k^{\text{p}} \right\}_{k=1}^{4}$, REMEMBER computes the cosine similarity for each class as:
\begin{equation*}
    s_k = \cos \left( \mathbf{v}_q^{\text{p}}, \mathbf{t}_k^{\text{p}} \right), \quad \text{for } k = 1, \dots, 4.
\end{equation*}

The predicted abnormality class is then obtained by selecting the anchor with the highest similarity:
\begin{equation*}
    \hat{y}_{\text{abn}} = \argmax_k s_k.
\end{equation*}

To estimate class probabilities, the similarity scores are normalized using a softmax function:
\begin{equation*}
    p_k = \frac{\exp(s_k)}{\sum_{j=1}^{4} \exp(s_j)}, \quad \text{for } k = 1, \dots, 4.
\end{equation*}

\subsection{Binary Dementia Classification}
We leverage the output of the abnormality type prediction to derive a binary dementia classification.
Specifically, if the predicted abnormality label is classified as \texttt{Normal}, we assign the dementia label $\hat{y}_{\text{dementia}} = 0$ (non-dementia);
otherwise, we assign $\hat{y}_{\text{dementia}} = 1$ (dementia). 

Let $s_{\text{normal}}$ denote the cosine similarity between the query embedding and the ``normal'' anchor, and $s_{\text{closest}}$ denote the similarity with the most similar abnormality anchor.
The probability of dementia is then defined as:
\begin{equation*}
    p_{\text{dementia}} = 
    \begin{cases}
        1 - s_{\text{normal}}, & \text{if the closest match is ``normal''} \\
        s_{\text{closest}}, & \text{otherwise}
    \end{cases}
\end{equation*}

\subsection{Dementia Type Classification}
To differentiate between major dementia categories, REMEMBER classifies each query image into one of three semantic classes based on textual anchors derived from the pseudo text modality \texttt{M.C} (see Appendix~\ref{appendix:data_MC}):
\begin{itemize}
    \item Non-dementia
    \item Alzheimer's Disease (AD)
    \item Other Dementia (e.g., Vascular Dementia, Lewy Body Dementia)
\end{itemize}

Following the same cosine similarity-based procedure, REMEMBER computes:
\begin{equation*}
    s_k = \cos \left( \mathbf{v}_q^{\text{p}}, \mathbf{t}_k^{\text{p}} \right), \quad \text{for } k = 1, 2, 3;
\end{equation*}
and derives: 
\begin{gather*}
    \hat{y}_{\text{type}} = \argmax_k s_k \\
    p_k = \frac{\exp(s_k)}{\sum{j=1}^{3} \exp(s_j)}
\end{gather*}

This task differs from abnormality classification in the nature of the labels -- it reflects diagnostic class rather than radiological presentation -- and uses a distinct anchor set specifically designed for clinical interpretation of dementia types.

\subsection{Dementia Severity Classification}
REMEMBER also performs fine-grained classification of dementia severity using text anchors derived from public datasets (modality \texttt{P}, see Appendix~\ref{appendix:data_P}).
The four predefined severity levels are:
\begin{itemize}
    \item Non-Demented
    \item Very Mild Demented
    \item Mild Demented
    \item Moderate Demented
\end{itemize}

Using the same formulation, we compute:
\begin{gather*}
    s_k = \cos \left( \mathbf{v}q^{\text{p}}, \mathbf{t}k^{\text{p}} \right), \quad k = 1, \dots, 4 \\
    \hat{y}_{\text{severity}} = \argmax_k s_k \\
    p_k = \frac{\exp(s_k)}{\sum{j=1}^{4} \exp(s_j)}
\end{gather*}

This task emphasizes sensitivity to early cognitive decline, distinguishing REMEMBER's utility in preclinical diagnosis and disease staging.

\subsection{Reference Case Retrieval}

In addition to classification via anchor matching, REMEMBER performs retrieval of reference cases from the MINDSet training corpus to support contextualized, evidence-based reasoning.
Each reference case in the corpus is embedded into the shared vision-text space during preprocessing.
At inference time, we compute the cosine similarity between the query image embedding $\mathbf{v}_q^{\text{p}}$ and all reference image embeddings $\{\mathbf{v}_i^{\text{p}}\}_{i=1}^{N}$.

The similarity for each candidate reference case is computed as:
\begin{equation*}
    \text{sim}_i = \cos \left( \mathbf{v}_q^{\text{p}}, \mathbf{v}_i^{\text{p}} \right)
\end{equation*}

The top-$k$ most similar reference cases are then selected based on the highest similarity scores.
Each retrieved case is associated with its corresponding clinical labels and pseudo text annotations (e.g., abnormality type, dementia label).
These references are used in downstream modules for evidence encoding, interpretability, and optionally report generation.

This retrieval process enables REMEMBER to ground its predictions in real, expert-annotated examples, reflecting how clinicians reason by comparing a new case with prior known cases.

\subsection{Output Format}

For each prediction task, REMEMBER provides both a diagnostic decision and contextual evidence to support interpretability. Specifically, the model outputs:

\begin{itemize}
    \item The predicted class label $\hat{y}$, derived via similarity to the most relevant text anchor.
    \item The softmax-normalized class probability vector $\mathbf{p}$, indicating confidence in each potential label.
    \item The top-$k$ retrieved reference cases from the MINDSet corpus, including:
    \begin{itemize}
        \item Corresponding similarity scores $\text{sim}_i$
        \item Clinical labels (e.g., abnormality type, dementia class)
        \item Associated pseudo text descriptions
    \end{itemize}
    \item Optionally, formatted reference-backed diagnostic reports that summarize the reasoning process in alignment with clinical documentation.
\end{itemize}

This structured output format supports transparent, explainable AI workflows by coupling predictive confidence with example-driven context.

% =================================================================
% APPENDIX Clinical Explanation Report Template
% =================================================================
\section{Clinical Explanation Report Template}
\label{appendix:report_template}

\subsection{Report Components}
Each REMEMBER report is designed to reflect the structure and style of clinical decision support outputs, offering a comprehensive view of the model's prediction and the rationale behind it.
The report integrates predicted outcomes, probabilistic confidence estimates, and retrieved reference cases, enabling clinicians to understand both what the model predicts and why it predicts it.
The main components include:

\textbf{Abnormality Type:} $\hat{y}_{\text{abn}}$ -- predicted structural abnormality label (e.g., MTL Atrophy, WMH, etc.).\\
\textbf{Abnormality Confidence:} $\mathbf{p}_{\text{abn}}$ -- softmax-normalized probabilities over the abnormality type classes.

\vspace{.5em}

\textbf{Dementia Diagnosis:} $\hat{y}_{\text{type}}$ -- predicted dementia subtype (e.g., AD, Other Dementia, Non-Dementia).\\
\textbf{Dementia Confidence:} $\mathbf{p}_{\text{type}}$ -- softmax-normalized probabilities over dementia type labels.

\vspace{.5em}

\textbf{Dementia Severity:} $\hat{y}_{\text{severity}}$ -- predicted stage of cognitive decline (e.g., Mild, Moderate).\\
\textbf{Severity Confidence:} $\mathbf{p}_{\text{severity}}$ -- softmax-normalized probabilities across severity levels.

\vspace{.5em}

\textbf{Evidence Table:} A ranked list of the top-$k$ retrieved reference cases, including:
\begin{itemize}
    \item \textbf{Similarity Score} ($\text{sim}_i$): Cosine similarity between the query and reference image.
    \item \textbf{Attention Weight} ($\alpha_i$): Learned attention score quantifying reference contribution to prediction.
    \item \textbf{Abnormality Type} ($y_i^{\text{abn}}$): Ground-truth abnormality label for reference case.
    \item \textbf{Dementia Label} ($y_i^{\text{dx}}$): Ground-truth dementia label for reference case.
    \item \textbf{Reference Description} ($\text{text}_{\text{desc},i}$): Human-readable radiology description.
\end{itemize}

\subsection{Example (Simplified)}

\noindent
\textbf{Abnormality Type:} MTL Atrophy \\
\textbf{Abnormality Confidence:} Normal: 3\%, MTL Atrophy: 91\%, WMH: 4\%, Other: 2\%

\vspace{.5em}

\noindent
\textbf{Dementia Diagnosis:} Alzheimer's Disease \\
\textbf{Dementia Confidence:} Non-Dementia: 6\%, AD: 89\%, Other Dementia: 5\%

\vspace{.5em}

\noindent
\textbf{Dementia Severity:} Mild Demented \\
\textbf{Severity Confidence:} Non-Demented: 7\%, Very Mild: 13\%, Mild: 72\%, Moderate: 8\%

\vspace{.5em}

\noindent
\textbf{Evidence Table:}

\begin{table}[!h]
\centering
\caption*{Top-3 Retrieved Reference Cases}
\resizebox{\linewidth}{!}{%
\begin{tabular}{|c|c|c|c|c|p{3.1cm}|}
\hline
\textbf{\#} & $\text{sim}_i$ & $\boldsymbol{\alpha}_i$ & \textbf{\begin{tabular}[c]{@{}c@{}}Abnor-\\ mality\end{tabular}} & \textbf{\begin{tabular}[c]{@{}c@{}}Dementia\\ label\end{tabular}} & \multicolumn{1}{c|}{\textbf{Reference description}} \\ \hline
1 & 0.94 & 0.57 & \begin{tabular}[c]{@{}c@{}}MTL\\ Atrophy\end{tabular} & AD & MRI shows severe hippocampal shrinkage and entorhinal cortex thinning, consistent with advanced medial temporal lobe atrophy. \\ \hline
2 & 0.89 & 0.36 & \begin{tabular}[c]{@{}c@{}}MTL\\ Atrophy\end{tabular} & AD & Evidence of moderate atrophy in the parahippocampal region and temporal horns enlargement. \\ \hline
3 & 0.85 & 0.07 & WMH & \begin{tabular}[c]{@{}c@{}}Other\\ Dementia\end{tabular} & MRI reveals scattered periventricular white matter hyperintensities, suggestive of small vessel ischemic changes. \\ \hline
\end{tabular}%
}
\label{tab:explanation_report}
\end{table}

This structured explanation provides traceability and interpretability, allowing clinicians to assess both the predicted outcome and the rationale behind it through real, contextually aligned reference cases.

% =================================================================
% APPENDIX Hyperparameters and Training Configurations
% =================================================================
\section{Hyperparameters and Training Configurations}
\label{appendix:hyperparameters}

This section details the training settings used to fine-tune the dual vision-text encoder and the downstream modules in REMEMBER.

\subsection{Vision-Text Contrastive Pretraining}

We follow a CLIP-style contrastive training setup~\cite{radford2021learning}, using the combined dataset of MINDSet and the public Alzheimer's MRI dataset for constructing image-text pairs (see Section~\ref{sec:dataset} and Appendix~\ref{appendix:text_modalities}).
The encoder weights are initialized from a pretrained BiomedCLIP model~\cite{zhang2024biomedclip} and further fine-tuned on our task-specific pairings.
\begin{itemize}
    \item \textbf{Batch size:} 16
    \item \textbf{Image resolution:} $224 \times 224$
    \item \textbf{Optimizer:} AdamW~\cite{loshchilov2017decoupled}
    \item \textbf{Learning rate:} $5 \times 10^{-5}$
    \item \textbf{Weight decay:} 0.2
    \item \textbf{Loss:} Symmetric cross-modal contrastive loss
    \item \textbf{Epochs:} 10
\end{itemize}

All training is conducted on a single NVIDIA TESLA P100 GPU via the Kaggle platform, with a total training time of approximately 2 hours.

\subsection{Text Encoders and Pseudo-modalities}
The text encoder uses a biomedical domain-adapted transformer from BiomedCLIP~\cite{zhang2024biomedclip}.
During training, each image is paired with four pseudo-text modalities from MINDSet (see Appendix~\ref{appendix:data_M}) and a single dementia severity description from the public dataset (see Appendix~\ref{appendix:data_P}), resulting in a rich and diverse set of image–text pairs for contrastive supervision.
Text inputs are tokenized using the PubMedBERT tokenizer~\cite{gu2021domain}.

\subsection{Evidence Encoding Module}
The evidence encoding module transforms each retrieved reference case into a compact embedding that captures multimodal context and similarity relevance.
Each evidence vector is constructed by concatenating both raw and similarity-weighted embeddings of four modalities: image, abnormality-type description, dementia label description, and original radiology-style description.
This yields an input feature of dimension $8 \times D$, where $D = 512$.

A single-layer feed-forward network is used to project the multimodal vector into a shared evidence space:
\begin{itemize}
    \item \textbf{Input dimension:} $8 \times 512$
    \item \textbf{Hidden layer:} None (single-layer projection)
    \item \textbf{Output dimension:} 512
    \item \textbf{Activation:} ReLU
    \item \textbf{Weight initialization:} Kaiming uniform~\cite{he2015delving}
\end{itemize}

\subsection{Attention-based Inference Head}

We use a single-head dot-product attention mechanism to compute attention over the encoded evidence matrix.
The final MLP for label prediction consists of one hidden layers with ReLU activations.
\begin{itemize}
    \item \textbf{Query projection:} Linear($512 \rightarrow 512$)
    \item \textbf{Evidence matrix:} $k = 3$ top reference cases
    \item \textbf{Final MLP:} $\left[ \mathbf{v}_q^{\text{p}}; \mathbf{z} \right] \rightarrow \text{MLP}(512 \rightarrow 256 \rightarrow C)$
    \item \textbf{Activation:} ReLU
    \item \textbf{Weight initialization:} Kaiming uniform
    \item \textbf{Loss function:} Cross-entropy
    \item \textbf{Batch size:} 4
    \item \textbf{Optimizer:} Adam~\cite{kingma2017adam}
    \item \textbf{Learning rate:} $5 \times 10^{-5}$
    \item \textbf{Early stopping:} Patience = 5 epochs (based on validation loss)
    \item \textbf{Max epochs:} 100
\end{itemize}

\subsection{Evaluation Settings}
During evaluation, we use frozen encoders and perform zero-shot and few-shot classification using similarity-based matching and attention-guided inference, without any further gradient updates.
Few-shot settings include $k=10$ examples per class unless otherwise specified.

\fi
\end{document}